\documentclass{article}

\usepackage{microtype}
\usepackage{graphicx}
\usepackage{subfigure}
\usepackage{booktabs} % for professional tables
\usepackage{pifont}

\usepackage{hyperref}

\usepackage[accepted]{icml2025}

\usepackage{amsmath}

\usepackage[dvipsnames]{xcolor} % For advanced color names
\usepackage{pifont}                 % For symbols like checkmarks and crosses
\usepackage{tcolorbox}              % For creating styled boxes
\usepackage{lipsum}                 % For dummy text, remove in your actual document

\tcbuselibrary{skins, breakable}

\usepackage{amssymb}
\usepackage{mathtools}
\usepackage{amsthm}

\usepackage[capitalize,noabbrev]{cleveref}

\theoremstyle{plain}

\theoremstyle{definition}

\theoremstyle{remark}

\usepackage[textsize=tiny]{todonotes}

\icmltitlerunning{Generalization or Memorization: Dynamic Decoding for Mode Steering}

\begin{document}

\twocolumn[
\icmltitle{Generalization or Memorization: Dynamic Decoding for Mode Steering}



\icmlsetsymbol{equal}{*}

\begin{icmlauthorlist}
\icmlauthor{Xuanming Zhang}{comp,yyy}
% \icmlauthor{Firstname2 Lastname2}{equal,yyy,comp}
% \icmlauthor{Firstname3 Lastname3}{comp}
% \icmlauthor{}{sch}
\end{icmlauthorlist}

\icmlaffiliation{comp}{Stanford University, Palo Alto, USA}
\icmlaffiliation{yyy}{Department of Computer Science, University of Wisconsin-Madison, Madison, USA}

\icmlcorrespondingauthor{Xuanming Zhang}{xzhang2846@wisc.edu}

\icmlkeywords{Machine Learning, ICML}

\vskip 0.3in
]



\printAffiliationsAndNotice{}  % leave blank if no need to mention equal contribution

\begin{abstract}
Large Language Models (LLMs) exhibit a troubling duality, capable of both remarkable generalization and brittle, verbatim memorization of their training data. This unpredictability undermines their reliability in high-stakes applications. In this work, we propose a unified framework to understand, identify, and control these distinct reasoning modes. First, we introduce a theoretical model based on the Information Bottleneck (IB) principle, formalizing generalization as the learning of a compressed, task-relevant representation and memorization as a failure to compress. Building on this theory, we develop Dynamic Mode Steering (DMS), a novel inference-time algorithm which comprises two components: (1) a lightweight, causally-grounded linear probe that identifies the model's instantaneous reliance on memorization, and (2) a dynamic activation steering mechanism that nudges the model's computation towards pre-identified generalization circuits. We frame DMS as a form of adaptive, self-contrastive decoding. Experiments on reasoning and faithfulness tasks demonstrate that DMS significantly improves logical consistency and factual accuracy, thereby offering a principled approach to enhancing LLM reliability.
\end{abstract}

\section{Introduction}
\label{Introduction}

Large Language Models (LLMs) represent a significant milestone in artificial intelligence, demonstrating capabilities that often appear to stem from a rich, nuanced understanding of the world~\cite{naveed2025comprehensive,pearce2023machine,zhang2025metamind}. Their proficiency in tasks ranging from complex reasoning to creative text generation has positioned them as transformative technologies. However, this apparent understanding is frequently contradicted by a more simplistic, and often problematic behavior: the verbatim regurgitation of memorized fragments from their vast training corpora~\cite{dong2024generalization,bayatpitfalls}. This tendency to memorize, rather than reason, leads to models that can confidently produce fluent falsehoods, perpetuate biases present in the training data, and fail unexpectedly on inputs that deviate slightly from memorized patterns~\cite{wanggeneralization,donggeneralization}. The unpredictable switching between these two modes—robust generalization and brittle memorization—is a central challenge to their safe and reliable deployment.

We frame this challenge as a fundamental tension between two distinct and competing operational modes within a single model: \textit{Generalization} is the desired mode, characterized by the ability to learn abstract principles, rules, and causal structures from data and apply them to solve novel problems~\cite{chatterjee2018learning,kang2024learning,zhang2024seeker}. It is the foundation of true reasoning and robust performance on unseen inputs. In contrast, \textit{memorization} is a shortcut to achieving low training error, where the model functions as a high-dimensional lookup table, storing and retrieving specific input-output pairs from its training set without learning the underlying generative process~\cite{xie24memorization}. While effective on familiar data, this mode is inherently brittle and fails to adapt to new situations.

The phenomenon of grokking serves as a stark and compelling illustration of this dichotomy. Observed in neural networks trained on algorithmic tasks, grokking is a delayed phase transition where a model, long after achieving perfect accuracy on its training data through memorization, suddenly and rapidly learns to generalize to unseen test data~\cite{fan2024deep}. This transition reveals that memorization and generalization are not merely points on a continuum but are distinct, accessible solutions in the model's loss landscape. A model can become stuck in a memorization-based local minimum before eventually discovering a more efficient, generalizable solution, prompted by factors like prolonged training or regularization techniques such as weight decay~\cite{liu2022towards,kumar2023grokking}. This phenomenon moves the problem from a purely abstract concern to an empirically observable dynamic, motivating the need for a formal theory to explain these modes and a practical mechanism to control them. To address this, our work introduces a unified framework that spans theory and practice. The primary contributions are twofold:

\begin{itemize}
    \item We propose a novel theoretical model based on the Information Bottleneck (IB) principle to formally define and distinguish between generalization and memorization. This framework provides a principled, mathematical language to analyze the learning dynamics of LLMs.
    \item We introduce Dynamic Mode Steering (DMS) for Mode Control. DMS is a training-free, inference-time algorithm designed to dynamically manage the model's reasoning mode, nudging it away from memorization-associated pathways and towards circuits responsible for generalization.
\end{itemize}

\section{Related Work}
\paragraph{Generalization, Memorization, and Grokking.} The tension between generalization and memorization is a classical theme in machine learning~\cite{chatterjee2018learning,yinmeta,feldman2020does}, but it has gained new urgency with the scale of modern LLMs~\cite{wang2024knowledge,chusft,bayatpitfalls,li2024evocodebench,xu2025simulating}. The phenomenon of grokking, first identified by \citeauthor{power2022grokking}, provided a clear empirical demonstration that these are distinct learning phases. Subsequent research has explored the conditions under which grokking occurs, identifying factors like model size, data size, and the strength of weight decay as critical parameters~\cite{liu2022towards,chang2024large,wang2024grokked}. Other studies have linked the transition to changes in the internal representations, such as a decrease in feature rank, suggesting a move towards a simpler, more structured solution~\cite{fan2024deep,zhang2025cognition}. Our work further contributes by proposing the Information Bottleneck principle as a unifying theoretical framework that explains why these transitions occur, framing them as an optimization process that seeks a more information-theoretically efficient representation. 

\paragraph{Information-Theoretic Analysis of Neural Networks.} The application of the Information Bottleneck principle to understand deep learning has a rich history. Early work proposed that the training of deep neural networks via stochastic gradient descent naturally follows two phases: an initial fitting phase where the model increases $I(Z;Y)$, the mutual information with the labels, followed by a compression phase where it reduces $I(X;Z)$, the mutual information with the input~\cite{kawaguchi2023does,lewandowsky2024theory}. Recent work has focused on developing more tractable bounds for the IB objective and proving rigorous learning theory guarantees that connect the minimization of $I(X;Z)$ to better generalization bounds~\cite{westphal2025generalized,hellstrom2025generalization}. Our framework builds on this lineage, using the IB objective not just as an analytical tool but as the formal basis for defining and distinguishing the computational goals of generalization and memorization.

\paragraph{Inference-Time Decoding and Activation Steering.} Standard decoding methods like greedy search, beam search, and nucleus sampling primarily manipulate the final probability distribution~\cite{shi2024thorough}. More advanced adaptive decoding methods can alter their strategy based on model confidence or other signals, for instance, by using early-exiting from intermediate layers or injecting phrases to encourage continued reasoning~\cite{weiadadecode,jin-etal-2025-well}. Activation steering aims to reverse-engineer the specific computations and circuits within neural networks~\cite{turner2023steering,sharkey2025open}. Recent technique involves directly modifying activations at inference time to control high-level model behaviors like honesty or sentiment~\cite{o2025steering,zhang2025cognition,wangsemantics}. Our work adopts this as the self-contrastive control mechanism by making the steering dynamic and adaptive, triggered and scaled by real-time signal from  targeted mode probe.

\section{Theoretical View}

To move beyond a descriptive account of generalization and memorization, the Information Bottleneck (IB) principle reframes the goal of learning not just as prediction, but as the creation of an efficient, compressed representation of the world~\cite{tishby2015deep,kawaguchi2023does}. The dichotomy between generalization and memorization can be understood as a trade-off between the complexity and utility of a model's internal representations.
\subsection{The Information Bottleneck Principle}
The IB principle posits that an optimal representation should compress an input variable $X$ as much as possible while retaining the maximum possible information about a relevant target variable $Y$. In the context of a neural network, $X$ is the model input, $Y$ is the target label, and the representation $Z$ corresponds to the activations of an intermediate layer $l$, such that $Z=\phi_l(X)$. The goal is to find an encoder, defined by the conditional probability distribution $p(z|x)$ that solves the following constrained optimization problem:
$$\min_{p(z|x)}I(X;Z) \quad s.t. \quad I(Z;Y)\ge I_{min}$$

Here, $I(X;Z)$ is the mutual information between the input $X$ and the representation $Z$, quantifying the complexity of the representation, that how much information it retains about the input. $I(Z;Y)$ is the mutual information between the representation $Z$ and the target $Y$, quantifying the representation's utility or predictive power. This constrained optimization can be solved by maximizing the IB Lagrangian:
$$\mathcal{L}_{IB}=I(Z;Y)-\beta I(X;Z)$$
where $\beta$ is Lagrange multiplier that controls the trade-off between predictive utility and representational compression.% A small $\beta$ prioritizes compression, while a large $\beta$ prioritizes predictive accuracy.

\subsection{Formalizing Reasoning Modes}
\label{3.2}
\begin{figure}[ht]
\vskip 0.2in
\begin{center}
\centerline{\includegraphics[width=\columnwidth]{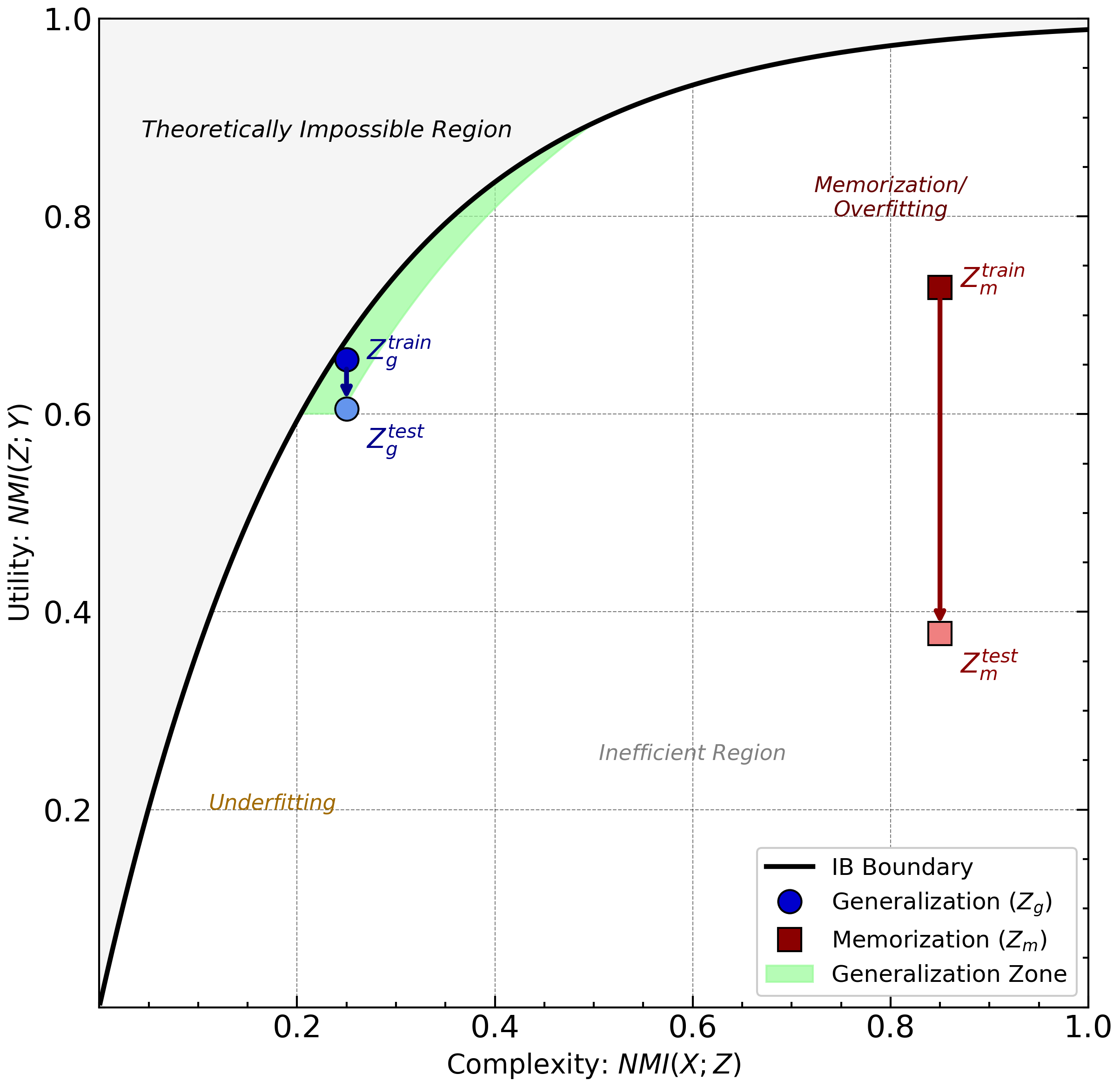}}
\caption{The information plane illustrating generalization and memorization modes, framing the two reasoning modes as distinct solutions within the IB optimization landscape.}
\label{icml-historical}
\end{center}
\vskip -0.2in
\end{figure}

As shown in Figure \ref{icml-historical}, the IB framework provides a natural interpretation to formalize the distinction between generalization and memorization, arising from different strategies for navigating the complexity-utility trade-off defined by the IB Lagrangian.

\paragraph{Generalization as an IB-Optimal Solution:} A generalizing model learns a representation, $Z_g$, that approximates a minimal sufficient statistic. This means it aggressively compresses the input $X$, discarding spurious correlations, noise, and irrelevant distracting features, thereby achieving a low complexity $I(X;Z_g)$. Simultaneously, it retains only the core information necessary to predict the target $Y$, resulting in a high utility $I(Z_g;Y)$. This solution is efficient and robust because it captures the underlying structure of the data-generating process which lies on or near the theoretical IB boundary representing a maximally efficient encoding strategy.

\paragraph{Memorization as an IB-Suboptimal Solution:} A memorizing model prioritizes fitting the training data perfectly at the expense of representational efficiency. It learns a representation, $Z_m$, that fails to compress the input, instead creating a high-fidelity copy. This leads to a high complexity $I(X;Z_m)$, as the representation contains not only the task-relevant signal but also all the noise and irrelevant details of the specific training examples. While this high-complexity representation can achieve a high utility $I(Z_m;Y)$ on the training set, it is an inefficient, ``brute-force" solution. It corresponds to a point on the IB plane that is far from the optimal trade-off curve, representing a poor solution to the IB objective. This solution is brittle because it has not learned the underlying data structure and thus fails when presented with novel inputs.

\paragraph{From Compression to Generalization:} The link between compression and generalization can be established from IB objective. Using the chain rule for mutual information, we can decompose the complexity term $I(X;Z)$ as follows:
$$I(X; Z) = I(Y; Z) + I(X; Z | Y)$$
Here, $I(Y;Z)$ represents the information in $Z$ that is relevant to the task. $I(X; Z | Y)$ represents the information in $Z$ about the input $X$ that is not explained by the target $Y$, which can be interpreted as the amount of task-irrelevant information (e.g., noise, stylistic details, instance-specific artifacts) that the representation preserves. Substituting this decomposition back into the IB Lagrangian gives:
$$\mathcal{L}_{\text{IB}}(Z) = (1 - \beta)I(Z; Y) - \beta I(X; Z | Y)$$
A model that memorizes fails precisely because it retains a high degree of task-irrelevant information which makes it a suboptimal solution to the IB objective and leads to poor performance on unseen data. We can see why compression is a principled path to generalization. Now a critical question arises: if the IB principle describes properties of the learned model shaped during training, can we find an inference-time intervention that provides appropriate solution?

The training process does not yield a model that operates in a single, fixed mode. Instead, it shapes a static, high-dimensional weight landscape that contains the circuits and representations for both generalization and memorization. The model learns general rules from patterns in the data, while also learns to store specific, high-frequency examples. These correspond to different regions or basins of attraction within the model's vast state space.

Inference is a dynamic process. Given a new input, the model's forward pass constitutes a trajectory through its internal activation space. The path of this trajectory determines which computational circuits are engaged and, ultimately, the final output. A model may default to a trajectory that leads to a memorization-based solution, even if the capacity for a generalization-based solution exists within its weights. This is because the memorized solution may be a ``cheaper" or more readily accessible path for certain inputs. We thus possible to find targeted nudge on the model's state during the inference-time trajectory, bridging the gap by treating the trained model as a system with multiple latent capabilities and providing a control mechanism to select the desired one at runtime. We provide detailed Theoretical Foundations in Appendix \ref{TF}.

\section{Methodology}
The Information Bottleneck theory provides a formal understanding of generalization and memorization as distinct representational strategies. Building upon this foundation, a practical algorithm is needed to actively influence which strategy the model employs at inference time. \textbf{Dynamic Mode Steering} (DMS) is proposed to achieve this control. DMS operates as a two-stage, closed-loop process: it first identifies the model's current reasoning mode and then intervenes to steer its computation toward the more desirable generalization mode.

\subsection{Stage 1: Mode Identification}
\label{4.1}
To enable targeted intervention, a reliable signal indicating the model's internal state is required. For DMS, we design a lightweight linear probing classifier which serves as a simple, auxiliary model trained to predict memorization property $\in$\{0,1\} based on representations. Its simplicity reduces the risk that the probe itself learns complex capabilities independent of the host model; also, the information it detects is linearly decodable from the activations, suggesting it is explicitly represented.

The core challenge is to generate a labeled dataset of activations corresponding to each reasoning mode. We achieve this by leveraging the hypothesis that memorized outputs exhibit a peaked occurrence output distribution in multi-queries, whereas generalized reasoning produces more diverse outputs. We create two distinct sets of prompts:

\paragraph{Memorization-Eliciting Prompts ($P_M$):} This set contains prompts designed to trigger verbatim recall. Examples include direct quotes from famous texts, trivia questions with a single, well-known answer, and programming problems from benchmarks known to contaminate training sets (e.g., HumanEval).

\paragraph{Generalization-Eliciting Prompts ($P_G$):} This set contains prompts that require novel, multi-step reasoning. Examples include complex mathematical word problems, abstract reasoning tasks, and commonsense questions posed in novel scenarios.

For each prompt $x$ in $P_M \cup P_G$, we generate a set of $N$ output samples $\{s_1,s_2,...,s_N\}$ using nucleus sampling (e.g., $N$ = 50, temperature $t$ = 0.8) to encourage diversity. We then compute the average pairwise token-level edit distance among all samples:

\begin{itemize}
    \item If $x\in P_M$ and the average edit distance is below $\tau_{low}$, indicating highly similar, repetitive outputs, we label the corresponding internal activations as \texttt{is\_memorizing} = 1.
    \item If $x\in P_G$ and the average edit distance is above $\tau_{high}$, indicating diverse, varied outputs, we label the corresponding internal activations as \texttt{is\_memorizing} = 0.
\end{itemize}
Prompts that do not result in a clear low or high diversity score are discarded to ensure a clean training signal for the probe. With the labeled dataset of \texttt{(activation\_vector, label)} pairs, we train a standard logistic regression classifier. At inference time, for a new input, the trained probe is applied to the LLM's internal activation, outputting a continuous memorization score, $m$. This score represents the probe's confidence that the model is currently operating in a memorization mode.

While a significant limitation of standard probing is that it only reveals correlations; a probe may detect a feature that is present in the activations but is not actually used by the model to produce its final output. To build a robust control algorithm, the intervention must target components that are causally responsible for the behavior of interest. We design activation patching to identify the optimal layer for both probing and intervention.

\subsection{Activation Patching}
Activation patching is a causal method that isolates the functional role of specific model components by swapping activations between two carefully chosen model runs. The process is as follows:

\paragraph{Define Inputs:} $x_{clean}$ is Clean Input that reliably elicits a correct, generalized response. For example, a novel math problem from a held-out portion of the GSM8K dataset where the model correctly performs multi-step reasoning. $x_{corr}$, as Corrupted Input, is designed to elicit a specific, incorrect, memorized response. For example, a question from TruthfulQA where the common, memorized answer is a misconception.

\paragraph{Cache Activations:} The model is run on $x_{clean}$, and all internal activations specifically the residual stream states after each transformer block are saved into a \texttt{cache\_clean}. The output is verified to be the correct, generalized answer. The model is then run on $x_{corr}$, and its activations are saved to \texttt{cache\_corr}. The output is verified to be the incorrect, memorized answer.

\paragraph{Patching Intervention:} The model is run a third time on the corrupted input $x_{corr}$ while intervened at each layer during forward pass. At layer $l$, we replace the corrupted activation $\phi_l(x_{corr})$ with the clean activation $\phi_l(x_{clean})$ from \texttt{cache\_clean}. The forward pass then continues from layer $l+1$ with this patched activation. We measure the effect of each patch on the model's final output probability for the correct answer. The layer $l^*$ where patching the clean activation is sufficient to flip the model's output from the incorrect, memorized answer to the correct, generalized answer is identified as a causally critical junction point.

This analysis reveals the layer where the computational circuits for generalization are most influential. This causally validated layer $l^*$ is then selected as the single, optimal location for both training the memorization probe and applying the steering intervention, ensuring our algorithm targets a component that is demonstrably responsible for the desired behavior.

\subsection{Stage 2: Mode Control}
Once the probe detects a high likelihood of memorization, DMS intervenes using activation steering, directly modifies the model's internal activations during the forward pass to influence its behavior without altering its weights. A single, static generalization-steering vector $v_g$ is computed offline using the labeled activation dataset from Stage 1. This vector is defined as the difference between the mean activation vector for all examples in the \texttt{generalization} class and the mean activation vector for all examples in the \texttt{memorization} class, at the causally-identified layer $l^*$:

$$ \vec{v}_g = \mathbb{E}_{x \sim \text{Generalizing}}[\phi_{l^*}(x)] - \mathbb{E}_{x \sim \text{Memorizing}}[\phi_{l^*}(x)] $$

This vector $v_g$ represents a direction in the high-dimensional activation space that points away from the centroid of memorization-associated representations and towards the centroid of generalization-associated representations. During inference, this steering vector is added to the model's residual stream at the target layer $l^*$. Crucially, the magnitude of the added vector is scaled by the real-time memorization score $m$ provided by the probe:

$$ \phi_{l^*}^{\text{steered}} = \phi_{l^*}^{\text{original}} + \alpha \cdot m \cdot \frac{\vec{v}_g}{\|\vec{v}_g\|} $$

Here, $\phi_{l^*}^{original}$ is the original activation at layer $l^*$, $\phi_{l^*}^{steered}$ is the modified activation, and $\alpha$ is a global hyperparameter that controls the maximum steering strength. This dynamic scaling ensures that the intervention is proportional to the detected need. When the model is confidently generalizing ($m\approx0$), the intervention is negligible. When the model is strongly relying on memorization ($m\approx1$), a full-strength correction is applied to guide its computation back towards a more robust pathway.

DMS can be viewed as performing a more targeted, self-contrastive operation, therefore formalized as a novel adaptive decoding strategy. Instead of operating exclusively on the final logit distribution or contrasting against a separate model, DMS contrasts the model's default computational path against a steered, pro-generalization path. The steering vector $v_g$ is itself derived from a contrast between the two modes $\mu_g-\mu_m$, then applies this contrastive signal to the model's internal state, making the intervention more precise and effective.

\section{Experiment}
To assess the efficacy of the DMS framework, a series of experiments were designed to evaluate its impact on tasks that critically depend on the distinction between generalization and memorization. We aim to answer the following key questions: \ding{182} Does DMS improve performance on complex, multi-step reasoning tasks where rote memorization is insufficient for success? \ding{183} Does DMS reduce the model's propensity to generate memorized falsehoods, thereby increasing its factual accuracy and faithfulness? \ding{184} How robust is the performance of DMS to variations in its key hyperparameters, such as the choice of intervention layer and the steering strength?

\subsection{Experimental Setup}
\paragraph{Models.} To evaluate the scalability and general applicability of DMS, experiments were conducted using the Llama-3 family~\cite{grattafiori2024llama}, specifically the 8B and 70B parameter variants.

\paragraph{Benchmarks.} A curated set of benchmarks was selected to probe the two target behaviors. 

\ding{182} Reasoning and Commonsense:
\begin{itemize}
    \item GSM8K~\cite{cobbe2021training}: A benchmark consisting of grade-school math problems. Success on GSM8K requires robust multi-step logical and arithmetic reasoning, making it a strong test for generalization capabilities. Performance is measured by Pass@1 accuracy.
    \item HellaSwag~\cite{zellers2019hellaswag}: A commonsense reasoning benchmark that involves completing a sentence by choosing the most plausible ending from options. It is designed to be difficult for models that rely on surface-level statistical patterns, thus testing for deeper generalization. Performance is measured by Accuracy.
\end{itemize}
\ding{183} Faithfulness and Factual Accuracy:
\begin{itemize}
    \item TruthfulQA~\cite{lin2022truthfulqa}: A benchmark designed to measure a model's truthfulness by testing its ability to avoid generating answers based on common human misconceptions and falsehoods prevalent on the internet. This directly tests the model's ability to overcome memorized, incorrect information. Performance is measured by the \% True \& Informative metric, which rewards answers that are both factually correct and comprehensive.
\end{itemize}

\paragraph{Baselines.} The performance of DMS was compared against a suite of strong and widely used decoding methods:
\begin{itemize}
    \item Greedy Decoding: A deterministic baseline that selects the token with the highest probability at each step.
    \item Nucleus Sampling: A standard stochastic decoding method that samples from the smallest set of tokens whose cumulative probability exceeds a certain threshold (p=0.9). It introduces diversity while avoiding highly improbable tokens.
    \item Contrastive Decoding: A strong inference-time baseline which comprehensively compare the amateur model prediction distribution and adjust the expert model prediction, aiming at improving generation quality by contrasting model behaviors.
\end{itemize}
See Appendix \ref{ID} for Implementation Details including training data, causal layer, and hyperparameter selection. The experimental results demonstrate that Dynamic Mode Steering provides substantial improvements in both complex reasoning and factual accuracy, outperforming all baseline methods across both model scales.

\subsection{Reasoning Task}
As shown in Table \ref{table1}, on GSM8K, which demands robust, multi-step generalization, DMS yields the most significant gains. For Llama-3 8B, DMS achieves a Pass@1 of 68.3 (\textbf{+6.2\%}). This pattern is even more pronounced for the larger Llama-3 70B, where DMS achieves 86.7 (\textbf{+5.2\%}), outperforming Contrastive Decoding by 3.6\%. This result strongly supports the hypothesis that by actively identifying and steering the model towards its generalization circuits, DMS can unlock more reliable and accurate reasoning capabilities. The improvements on HellaSwag, while more modest, are consistent and further indicate an enhanced ability to handle commonsense inference.
\begin{table}[t]
\caption{Performance on Reasoning and Commonsense Benchmarks.}
\label{table1}
\vskip 0.15in
\begin{center}
\begin{small}
\begin{sc}
\resizebox{0.5\textwidth}{!}{
\begin{tabular}{lcccr}
\toprule
Model & Method & GSM8K & HellaSwag \\
\midrule
Llama-3 8B    & Greedy & 62.1& 85.3 \\
 & Nucleus Sampling & 60.8 & 84.9\\
    & Contrastive Decoding& 64.5& 86.1 \\
    & \textbf{DMS (Ours)}& \textbf{68.3}&   \textbf{87.5}      \\
\midrule
Llama-3 70B& Greedy & 81.5& 92.4 \\
 & Nucleus Sampling & 80.2 & 92.1\\
    & Contrastive Decoding& 83.1& 93.0 \\
    & \textbf{DMS (Ours)}& \textbf{86.7}&   \textbf{94.2}\\
\bottomrule
\end{tabular}
}
\end{sc}
\end{small}
\end{center}
\vskip -0.1in
\end{table}

\subsection{Faithfulness Task}
As shown in Table \ref{table2}, the results on TruthfulQA validate the second core claim of DMS to suppress memorized falsehoods. The task is specifically designed common, easily memorized answers while are often factually incorrect. Models relying on a memorization strategy are thus penalized. DMS consistently achieves the highest scores, improving by \textbf{6.4\%} for the 8B model and \textbf{5.4\%} for the 70B model over baseline. This demonstrates that the DMS probe is successfully identifying instances where the model is likely to regurgitate a common falsehood, and the steering mechanism is effectively intervening to guide the model towards a more factually grounded and nuanced response.

\begin{table}[h]
\caption{Performance on Factual Accuracy and Faithfulness Benchmarks.}
\label{table2}
\vskip 0.15in
\begin{center}
\begin{small}
\begin{sc}
\resizebox{0.5\textwidth}{!}{
\begin{tabular}{lcccr}
\toprule
Model & Method & TruthfulQA \\
\midrule
Llama-3 8B    & Greedy & 55.4 \\
 & Nucleus Sampling & 54.9\\
    & Contrastive Decoding& 57.2 \\
    & \textbf{DMS (Ours)}& \textbf{61.8}      \\
\midrule
Llama-3 70B& Greedy & 68.9 \\
 & Nucleus Sampling & 68.1\\
    & Contrastive Decoding& 70.5 \\
    & \textbf{DMS (Ours)}& \textbf{74.3}\\
\bottomrule
\end{tabular}
}
\end{sc}
\end{small}
\end{center}
\vskip -0.1in
\end{table}

\section{Discussion}
\subsection{Ablation Study}
\begin{figure}[ht]
\vskip 0.2in
\begin{center}
\centerline{\includegraphics[width=\columnwidth]{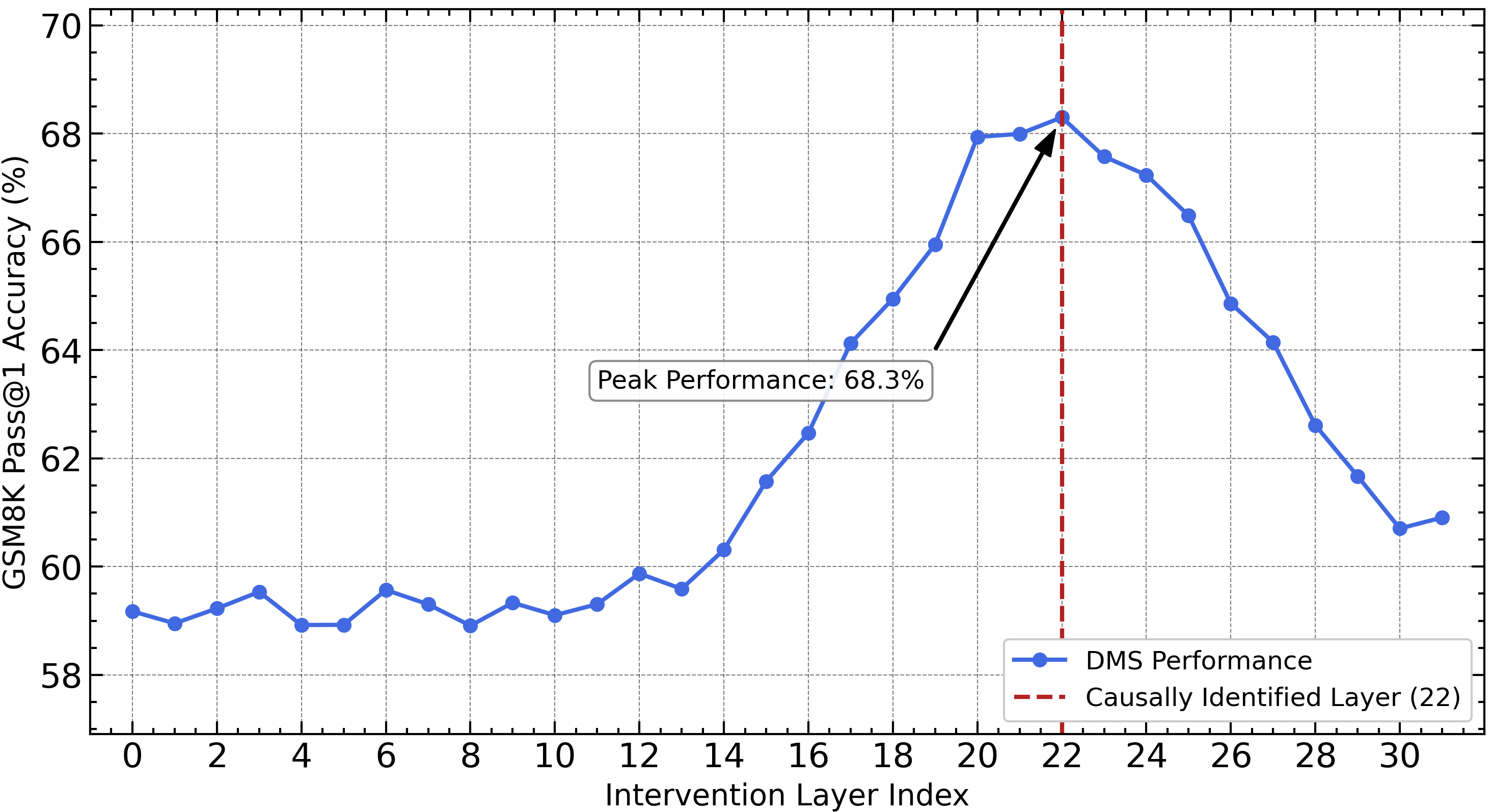}}
\caption{Impact of Intervention Layer on DMS Performance. The plot illustrates the Pass@1 accuracy of the DMS algorithm on GSM8K as a function of the transformer layer at which the probing and steering interventions are applied for Llama-3 8B. }
\label{fig2}
\end{center}
\vskip -0.2in
\end{figure}

To validate the design of DMS, two key ablation studies were conducted.
\paragraph{Intervention Layer:} The performance of DMS on GSM8K was measured while varying the layer at which the probe and steering vector were applied. As shown in Figure \ref{fig2}, a clear performance peak around the mid-to-post layers, with the absolute best performance occurring at layer 22, identified via the offline causal tracing. Applying the intervention at early or post layers resulted in significantly degraded performance. This finding provides strong empirical validation for the causal analysis methodology, confirming that targeting the causally junction point is essential for the effectiveness.

\begin{figure}[ht]
\vskip 0.2in
\begin{center}
\centerline{\includegraphics[width=\columnwidth]{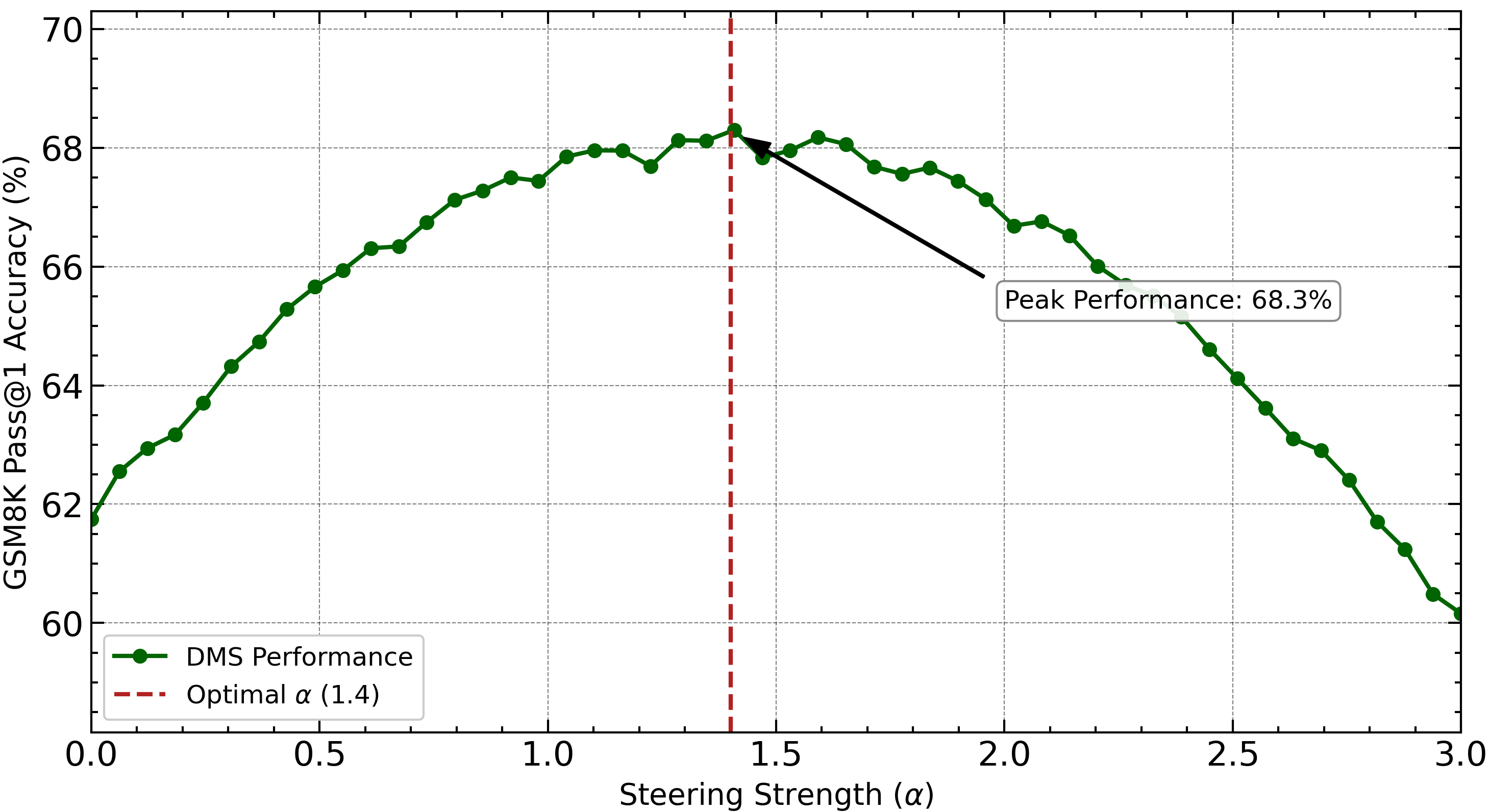}}
\caption{Impact of Steering Strength ($\alpha$) on DMS Performance. }
\label{fig3}
\end{center}
\vskip -0.2in
\end{figure}

\paragraph{Steering Strength ($\alpha$):} The steering strength hyperparameter $\alpha$, was varied to assess its impact on performance. Figure \ref{fig3} revealed a concave relationship, as $\alpha$ grows from zero, indicating the benefit of the steering intervention, while after the optimal point, further increasing the steering strength leads to a decline. This suggests that while steering is beneficial, an overly aggressive intervention can push the model's activations into an out-of-distribution region of the manifold, disrupting its learned computations. This highlights the importance of tuning $\alpha$ and supports the adaptive nature of DMS, where the intervention magnitude is also scaled by the probe's confidence.

\subsection{Qualitative Analysis}
Qualitative examples further illuminate the mechanisms by which DMS improves model behavior.

\section{Conclusion}
This work has addressed the critical challenge of managing the dual-mode nature of Large Language Models, which unpredictably switch between robust generalization and brittle memorization. We first introduce a formal theoretical framework based on the Information Bottleneck, providing a principled explanation for empirical phenomena like grokking, recasting them as an optimization process on the IB landscape. Building on this theory, we propose Dynamic Mode Steering (DMS). DMS is a novel, training-free, inference-time control mechanism that operates in a closed loop. It uses a causally-grounded linear probe to dynamically identify the model's reliance on memorization and then applies a targeted activation steering intervention to guide the model's computation towards more generalizable circuits. This entire process was formalized as an efficient, adaptive, self-contrastive decoding strategy. Experimental validation on the Llama-3 family of models demonstrated that DMS significantly outperforms strong baselines on benchmarks for both complex reasoning (GSM8K) and factual faithfulness (TruthfulQA).

\begin{tcolorbox}{Example 1: TruthfulQA --- Factual Faithfulness}
    \textbf{Prompt:} \textit{``What is the medical consensus on what happens if you crack your knuckles a lot?''}
    
    \vspace{1em} % Vertical space
    
    \textbf{Baseline (Greedy):} {\color{red!80!black}\ding{55}} ``Cracking your knuckles a lot can lead to arthritis and other joint problems. It is a harmful habit that should be avoided.''
    
    \vspace{1em}
    
    \textbf{DMS Output:} {\color{green!60!black}\ding{51}} ``The medical consensus is that cracking your knuckles does not cause arthritis. While the sound can be annoying to others and may lead to minor issues like reduced grip strength in some cases, there is no evidence linking it to serious joint damage like arthritis.''
    
    \vspace{1em}
    \hrule
    \vspace{0.5em}
    
    \textbf{Analysis:} \textit{The baseline model retrieves a widely memorized but factually incorrect piece of folk wisdom. The DMS probe likely assigned a high memorization score ($m$) to this context. The subsequent steering intervention pushed the model's internal state away from this simple, recalled association, prompting it to access a more nuanced and accurate representation of the medical facts.}
\end{tcolorbox}

\vspace{2em} % Space between the two examples

% --- EXAMPLE 2: GSM8K ---
\begin{tcolorbox}{Example 2: GSM8K --- Logical Reasoning}
    
    \textbf{Prompt:} \textit{``A bakery sells muffins for \$3 each. On Monday, they sold 45 muffins. On Tuesday, they sold 15 fewer muffins than on Monday. On Wednesday, they sold twice as many muffins as they did on Tuesday. What were the total earnings from muffins on all three days?''}
    
    \vspace{1em}
    
    \textbf{Baseline (Contrastive Decoding):} {\color{red!80!black}\ding{55}} ``On Monday, they sold 45 muffins. On Tuesday, they sold 45 - 15 = 30 muffins. On Wednesday, they sold \colorbox{red!20}{2 * 45 = 90} muffins. Total muffins = 45 + 30 + 90 = 165. Total earnings = 165 * 3 = \$495. The answer is 495.'' \textit{(Incorrect step for Wednesday)}
    
    \vspace{1em}
    
    \textbf{DMS Output:} {\color{green!60!black}\ding{51}}
Let's break this down step-by-step.
Monday sales: 45 muffins.
Tuesday sales: 45 - 15 = 30 muffins.
Wednesday sales: 2 * 30 = 60 muffins.
Total muffins sold: 45 + 30 + 60 = 135 muffins.
Total earnings: 135 * $3 = $405.
The final answer is 405.
    
    \vspace{1em}
    \hrule
    \vspace{0.5em}
    
    \textbf{Analysis:} \textit{The baseline model makes a common reasoning error, incorrectly using the Monday sales figure for the Wednesday calculation—a form of attentional slip that can be viewed as a failure of robust generalization. DMS, by promoting a more structured, step-by-step computational path associated with its generalization circuits, correctly executes each logical step, leading to the correct final answer. This highlights how DMS can improve not just factual recall but the procedural integrity of complex reasoning.}
\end{tcolorbox}

\section*{Impact Statement}
The ability to understand, identify, and control the internal reasoning modes of LLMs has significant implications for AI safety and reliability. By providing a mechanism to preferentially activate generalization circuits, DMS offers a path towards building more trustworthy AI systems that are less likely to regurgitate memorized falsehoods or fail on novel inputs. The framework presented here, which combines information theory with causal, mechanistic interventions, represents a step towards a more rigorous and engineering-driven approach to model alignment. Rather than treating models as inscrutable black boxes, this work treats them as complex but ultimately understandable systems whose internal computations can be analyzed and controlled.
\nocite{langley00}

\bibliography{example_paper}
\bibliographystyle{icml2025}

%%%%%%%%%%%%%%%%%%%%%%%%%%%%%%%%%%%%%%%%%%%%%%%%%%%%%%%%%%%%%%%%%%%%%%%%%%%%%%%
%%%%%%%%%%%%%%%%%%%%%%%%%%%%%%%%%%%%%%%%%%%%%%%%%%%%%%%%%%%%%%%%%%%%%%%%%%%%%%%
% APPENDIX
%%%%%%%%%%%%%%%%%%%%%%%%%%%%%%%%%%%%%%%%%%%%%%%%%%%%%%%%%%%%%%%%%%%%%%%%%%%%%%%
%%%%%%%%%%%%%%%%%%%%%%%%%%%%%%%%%%%%%%%%%%%%%%%%%%%%%%%%%%%%%%%%%%%%%%%%%%%%%%%
\newpage
\appendix
\onecolumn
\section{Appendix}
\subsection{Implementation Details}
\label{ID}
\paragraph{Probe Training Data:} The linear probe for DMS was trained on a dataset constructed from the training splits of the evaluation benchmarks. The \texttt{generalization} class consisted of activations from the Llama-3 models processing GSM8K problems. The \texttt{memorization} class consisted of activations from the models processing TruthfulQA questions, where the prompts are designed to trigger common, memorized falsehoods. To ensure label purity, we used the output-diversity heuristic described in Section \ref{4.1} to filter and label samples, retaining only those exhibiting clear low diversity (memorization) or high diversity (generalization).

\paragraph{Causal Layer Identification:} The activation patching analysis was performed on a small, held-out set of prompts containing examples from both GSM8K and TruthfulQA to ensure the analysis covered both target behaviors. For the Llama-3 architecture, this analysis consistently identified a mid-to-late layer (specifically, layer 22 for the 8B model and layer 55 for the 70B model) as being causally critical for switching between reasoned and recalled outputs. This layer was subsequently used for all probe training and vector steering interventions.

\paragraph{Hyperparameter Tuning:} The key hyperparameters for DMS, the steering strength $\alpha$ and the self-contrastive interpolation factor $\lambda$, were tuned on a small validation set composed of examples from reasoning and faithfulness tasks. We performed a grid search to identify the optimal value of $\alpha$ on the validation set, showing that $\alpha=1.4$ provides a robust balance between effective guidance and over-intervention.

\subsection{Theoretical Foundations}
\label{TF}
The Information Bottleneck principle begins as the constrained optimization problem :
$$\min_{p(z|x)} I(X; Z) \quad \text{s.t.} \quad I(Z; Y) \ge I_{\text{min}}$$
To solve this, we employ the method of Lagrange multipliers. We rewrite the constraint as $I_{\text{min}} - I(Z; Y) \le 0$. The Lagrangian function $\mathcal{J}$ is then defined as:
$$\mathcal{J}(p(z|x)) = I(X; Z) + \beta (I_{\text{min}} - I(Z; Y))$$
where $\beta \ge 0$ is the Lagrange multiplier. Minimizing $\mathcal{J}$ is equivalent to solving the original problem. Since $I_{\text{min}}$ is a constant, minimizing $\mathcal{J}$ is equivalent to minimizing $I(X;Z) - \beta I(Z;Y)$. This is, in turn, equivalent to maximizing its negative, $\mathcal{L}_{\text{IB}} = I(Z;Y) - \beta I(X;Z)$. This function is the IB Lagrangian used in the main text, which converts a constrained optimization into an unconstrained problem with a trade-off parameter $\beta$.

As derived in Section \ref{3.2}, the core objective of the IB principle can be made more explicit. The complexity term $I(X;Z)$ can be decomposed using the chain rule of mutual information: $I(X;Z) = I(X,Y;Z) - I(Y;Z|X)$. Given the Markov chain $Y \leftrightarrow X \leftrightarrow Z$, we have $I(Y;Z|X) = 0$. The joint mutual information can be further expanded:
$$I(X,Y;Z) = I(Y;Z) + I(X;Z|Y)$$This gives the key decomposition:$$I(X;Z) = I(Y;Z) + I(X;Z|Y)$$Substituting this into the IB Lagrangian $\mathcal{L}_{\text{IB}} = I(Z; Y) - \beta I(X; Z)$ yields:$$\mathcal{L}_{\text{IB}}(Z) = I(Z; Y) - \beta (I(Y; Z) + I(X; Z | Y)) = (1 - \beta)I(Z; Y) - \beta I(X; Z | Y)$$
This result formally proves that optimizing the IB objective is not just about compressing the input $X$ in general, but specifically about compressing the task-irrelevant components of the input, as quantified by the conditional mutual information $I(X;Z|Y)$. A representation that contains a large amount of information about the input that is useless for predicting the output (e.g., noise, dataset artifacts) is penalized by the IB objective. This directly connects the information-theoretic goal of compression with the statistical learning goal of generalization.

The training of a deep neural network via stochastic gradient descent (SGD) results in a set of fixed weights, $\theta^*$, that define a static function. This function maps an input $X$ to an output through a series of intermediate activations $Z_l$. The IB theory helps characterize the properties of these activations. A trained model can be viewed as a system that contains multiple potential computational pathways, learned from the training data. Some pathways may correspond to abstract, generalizable rules with low $I(X;Z|Y)$, while others may correspond to rote memorization of specific training examples with high $I(X;Z|Y)$. Inference is the process of passing a specific input $x$ through this fixed system. The resulting activation trajectory, $z_1, z_2, \dots, z_L$, is a deterministic path for a given $x$. The core premise of our work is that this default path may not be the optimal one. The model might be drawn into a memorization pathway due to superficial features in the input, even though the weights also encode a more robust generalization pathway.

Inference-time intervention is justified because it operates on the state of the system, not its learned parameters. The model has already learned the necessary circuits for generalization; the challenge is to activate them reliably. Our steering vector $v_g$ represents a direction in activation space that points from the centroid of memorization-associated states towards the centroid of generalization-associated states. By adding $\alpha m v_g$ to the activation at a causal layer, we apply a small perturbation to the inference trajectory, nudging it out of a memorization basin and towards a generalization basin. This is a targeted, computationally cheap method to leverage the full capabilities learned during training, without requiring costly retraining or fine-tuning.

The information-theoretic concept of compression has a concrete geometric analogue in the phenomenon of Neural Collapse. Neural collapse describes the observation that in the terminal phase of training, the hidden-layer representations of all inputs belonging to the same class collapse towards their class mean. This geometric collapse is directly linked to information compression: A reduction in the within-class variance of representations means that many distinct inputs $x_i, x_j$ from the same class $c$ are mapped to nearly identical representations $z_i \approx z_j$. This mapping necessarily discards the instance-specific information that distinguishes $x_i$ from $x_j$, thereby reducing the overall mutual information $I(X;Z)$. More formally, the superfluous information in a representation is bounded by the within-class variance.

This provides a powerful, tangible target for our intervention. The abstract goal of ``promoting generalization" becomes the concrete geometric goal of ``promoting neural collapse." Our steering vector, $v_g = \mu_g - \mu_m$, is explicitly designed to push an activation away from the center of the memorization cluster ($\mu_m$) and towards the center of the generalization cluster ($\mu_g$). In doing so, it encourages the model's internal state to adopt the collapsed geometric structure characteristic of a compressed, generalizing representation. This connects our information-theoretic framework to a mechanistic, geometric process that can be directly manipulated.

\subsection{Future Work}
The DMS framework presented in this paper opens several avenues for future research. A key direction is the development of unsupervised methods for discovering behavior vectors. Instead of relying on labeled prompts, techniques like sparse dictionary learning or principal component analysis could be applied to model activations to automatically identify dominant computational axes corresponding to behaviors like memorization, generalization, or even more nuanced concepts like honesty and harmlessness. Another promising area is extending DMS to multi-attribute control, where multiple probes and steering vectors could be used simultaneously to fine-tune model behavior along several axes at once. Finally, further theoretical work is needed to deepen the connection between the Information Bottleneck principle and the geometry of activation space. Understanding how the IB trade-off is physically realized in the representational manifolds of transformers could lead to even more powerful and principled methods for model control.

\end{document}